\pdfoutput=1
\documentclass{article}
\usepackage{fullpage}
\usepackage[authoryear,round,sort]{natbib}

\usepackage{extarrows}
\usepackage{wrapfig}
\usepackage{amsmath,amssymb,mathtools}
\usepackage{bm}
\usepackage{amsthm}
\usepackage{graphicx,subfig,color,wrapfig,epsfig,epstopdf}
\usepackage{enumitem}
\usepackage{booktabs}
\usepackage{algorithm,algorithmic}

\renewcommand{\paragraph}[1]{\noindent\textbf{#1}}

\usepackage[utf8]{inputenc} % allow utf-8 input
\usepackage[T1]{fontenc}    % use 8-bit T1 fonts
\usepackage{hyperref}       % hyperlinks
\definecolor{darkblue}{rgb}{0, 0, 0.5}
  \hypersetup{colorlinks=true,citecolor=darkblue, linkcolor=darkblue, urlcolor=darkblue}
\usepackage{url}            % simple URL typesetting
\usepackage{booktabs}       % professional-quality tables
\usepackage{amsfonts}       % blackboard math symbols
\usepackage{mathrsfs}
\usepackage{nicefrac}       % compact symbols for 1/2, etc.
\usepackage{microtype}      % microtypography

\usepackage{color}

\usepackage[sc]{mathpazo} % Use the Palatino font
\usepackage[T1]{fontenc}  % Use 8-bit encoding that has 256 glyphs
\usepackage{microtype}    % Slightly tweak font spacing for aesthetics
\usepackage{abstract}     % Allows abstract customization
\allowdisplaybreaks[4]

\usepackage{amsmath,amssymb,mathtools}
\usepackage{bm}
\usepackage{amsthm}
\usepackage{graphicx,subfig,color,wrapfig,epsfig,epstopdf}
\usepackage{enumitem}
\usepackage{booktabs}
\usepackage{algorithm,algorithmic}
\title{Strictly Breadth-First AMR Parsing}
\date{}

\usepackage{authblk}
\author[]{Chen Yu\thanks{\texttt{cyu28@ur.rochester.edu}} }
\author{Daniel Gildea\thanks{\texttt{gildea@cs.rochester.edu}}}
\affil{Department of Computer Science, University of Rochester}

\begin{document}

\maketitle

\begin{abstract}
AMR parsing is the task that maps a sentence to an AMR semantic graph automatically. We focus on the breadth-first strategy of this task, which was proposed recently and achieved better performance than other strategies. However, current models under this strategy only \emph{encourage} the model to produce the AMR graph in breadth-first order, but \emph{cannot guarantee} this. To solve this problem, we propose a new architecture that \emph{guarantees} that the parsing will strictly follow the breadth-first order. In each parsing step, we introduce a \textbf{focused parent} vertex and use this vertex to guide the generation. With the help of this new architecture and some other improvements in the sentence and graph encoder, our model obtains better performance on both the AMR 1.0 and 2.0 dataset.
\end{abstract}

%This architecture is designed to mimic how humans produce an AMR in a semantic-based way. With help of the parent information, this model is able to produce its children more precisely. We remove the complex re-categorization process which was used in most previous work. We argue that with increase of dataset size, the benefit of re-categorization process will disappear, and we demonstrate it in our experiment. Our model obtains a new SOTA performance among models without re-categorization process. Under the Nowiki-Smatch metric, our model becomes the SOTA even comparing with models with re-categorization process on AMR 2.0 dataset.

\section{Introduction} \label{sec:intro}
\begin{figure*}[tbp]
    \centering
    \includegraphics[width=0.5\textwidth]{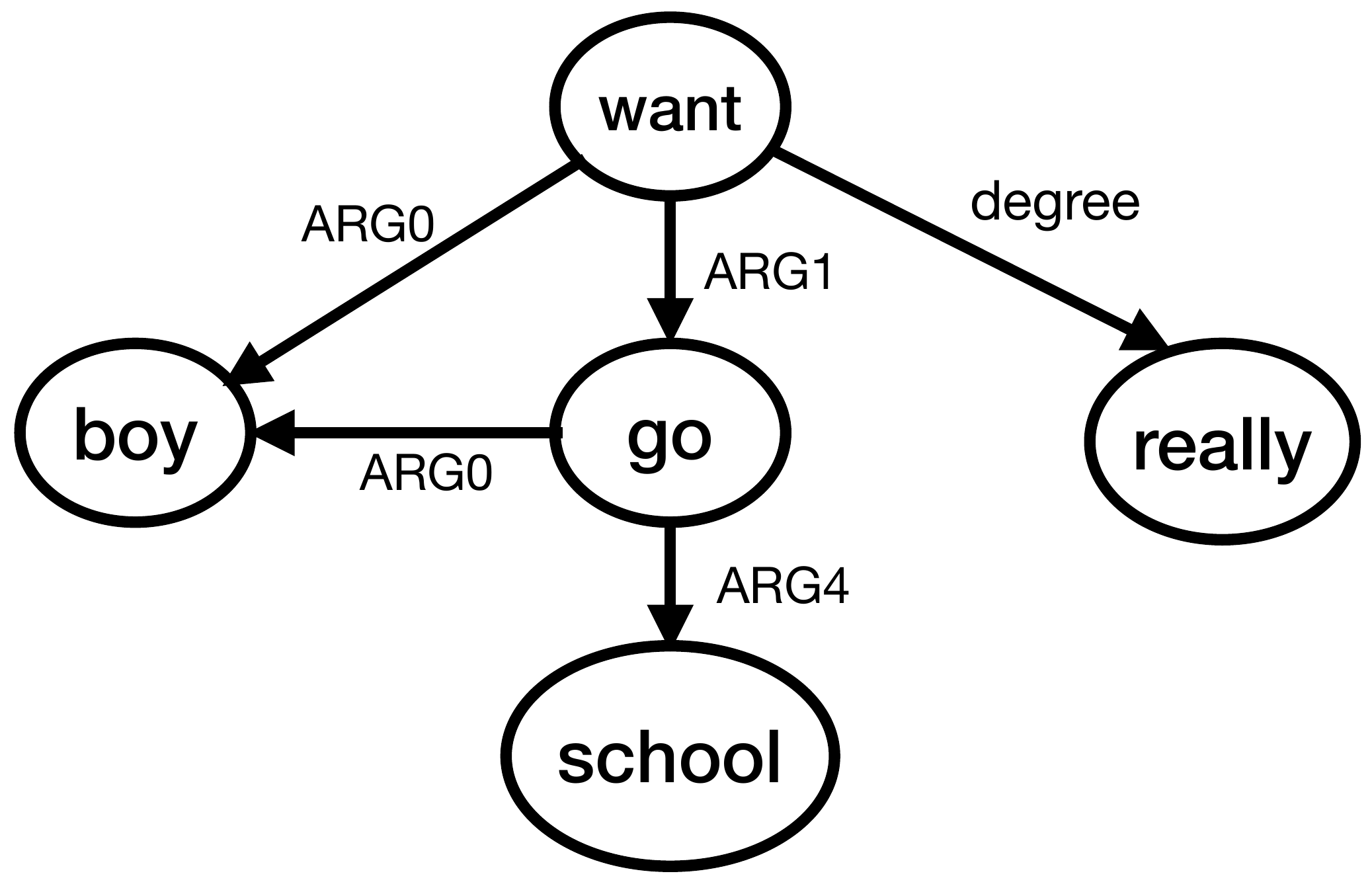}
    \caption{AMR Graph for the Sentence: \textit{The boy really wants to go to school.}}
    \label{fig:example}
\end{figure*}

Abstract Meaning Representation (AMR) \citep{banarescu2013abstract} is a graph that encodes the semantic meaning of a sentence. In Figure~\ref{fig:example}, we show the AMR of the sentence: \textit{The boy really wants to go to school}. The vertices in AMR represent concepts in the sentence and edges represent the relation between two concepts. From the root vertex, which is usually the key concept in the sentence, the graph gradually elaborates the details of the sentence as the depth of the graph increases. AMR has been widely used in many NLP tasks \citep{liu2015toward, hardy2018guided, mitra2016addressing}. 

AMR parsing is the task that maps a sentence to an AMR semantic graph automatically. A graph is a complex data structure which is composed of multiple vertices and edges. To produce a graph, one must determine the order of producing these vertices and edges. There are roughly four kinds of orders in previous work:

\begin{itemize}
    \item \textbf{Two-Stage Order} \citep{flanigan-acl14, lyu-acl18, zhang-etal-2019-amr, zhou-etal-2020-amr}: first produce vertices, and produce edges after that.
    \item \textbf{Transition-Based Order} \citep{ DamonteCS16, ballesteros2017amr, guo-lu-2018-better, wang-xue-2017-getting, naseem2019rewarding}: process the sentence from left to right, and produce vertices and edges based on the current focused word.
    \item \textbf{Depth-first Based Order} \citep{konstas-acl17, van2017neural,  peng-eacl17, peng-acl18, zhang-etal-2019-broad}: produce vertices and edges based on depth-first order.
    \item \textbf{Breadth-first Based Order} \citep{cai2019core, cai-lam-2020-amr}: produce vertices based on breadth-first order, and produce edges between the new vertex and existing vertices when producing each vertex.
\end{itemize}

We focus on the last type --- \emph{breath-first based order}, which was proposed by \citeauthor{cai-lam-2020-amr} and achieves better performance than other strategies. In this order, vertices and edges that are close to the root vertex are produced first. In AMR, these vertices usually capture the core semantics of a sentence. Therefore, models in this order pay more attention to the core semantics, which is one reason that they usually show better performance.

\begin{figure*}[tbp]
    \centering
    \includegraphics[width=0.9\textwidth]{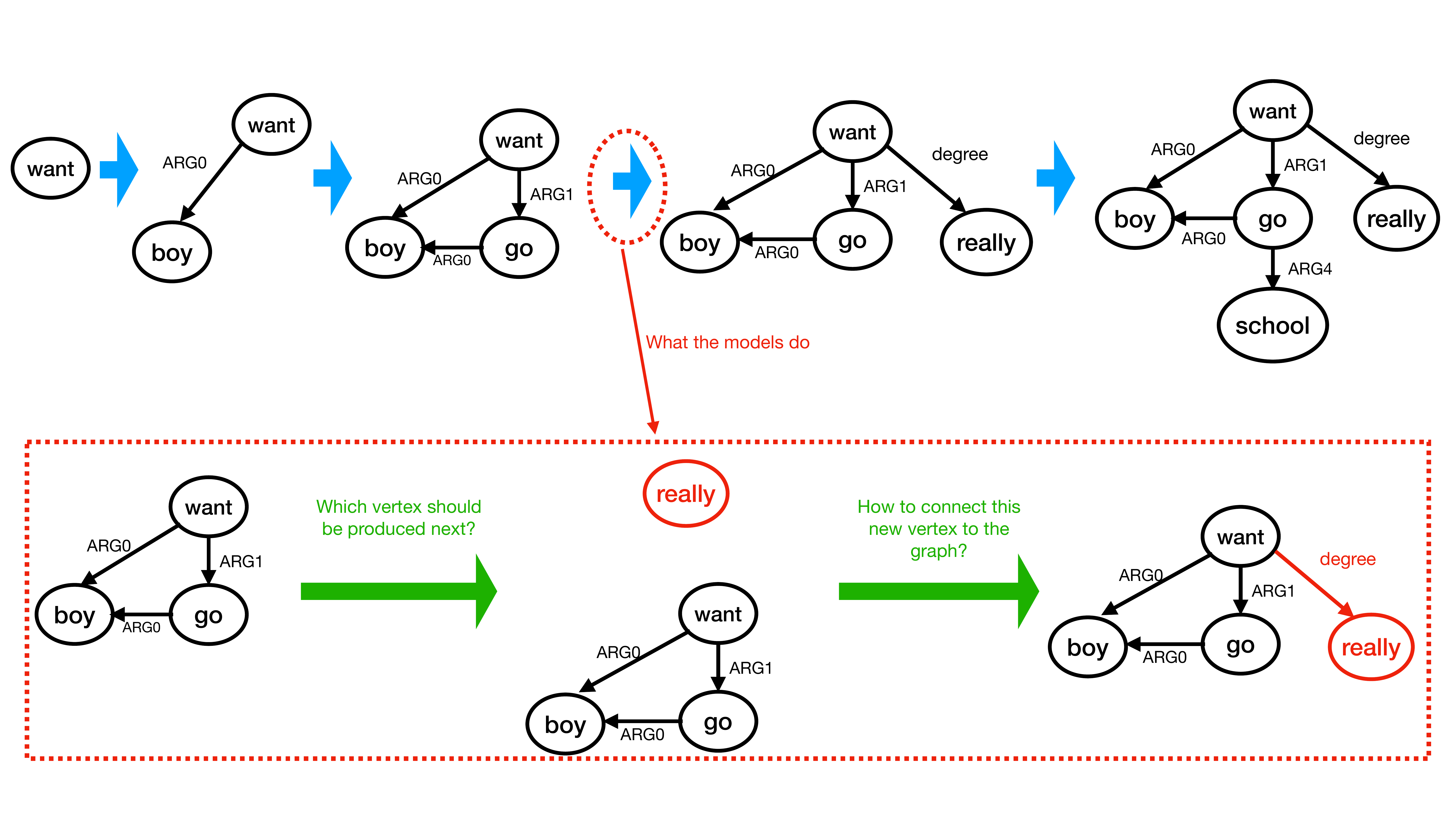}
    \caption{How previous methods generate the AMR graph in the breadth-first way for the sentence: \textit{The boy really wants to go to school.}}
    \label{fig:previous_models}
\end{figure*}

\begin{figure*}[tbp]
    \centering
    \includegraphics[width=0.9\textwidth]{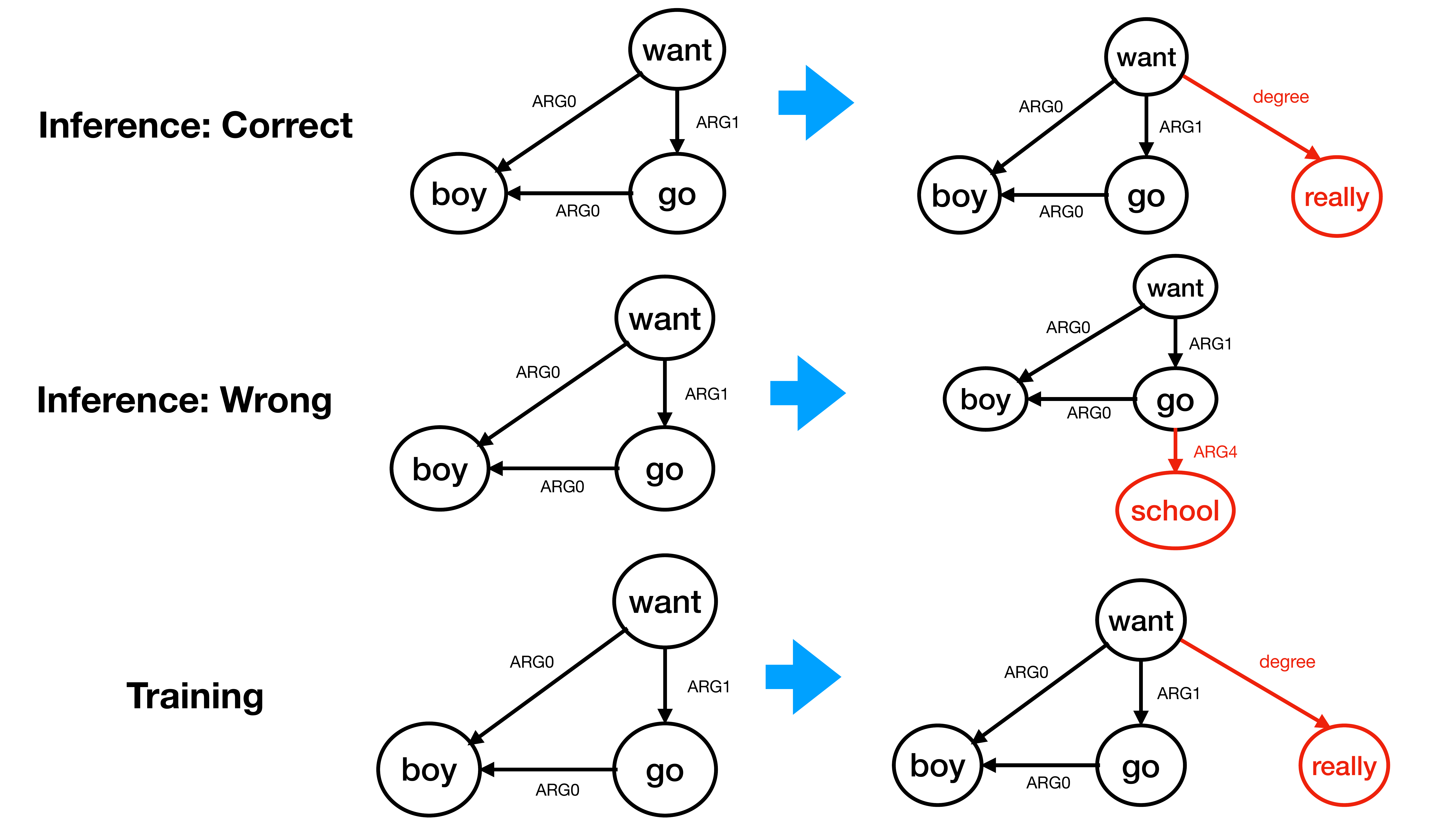}
    \caption{No guarantee of the breadth-first order for inference. Training process encourages the breadth-first order by always choosing the gold next-child in this order.}
    \label{fig:previous_mistake}
\end{figure*}

However, the existing breadth-first based methods \citep{cai2019core, cai-lam-2020-amr} are not truly breadth-first. They only \emph{encourage} the model to produce the AMR graph in the breadth-first way, but \emph{cannot guarantee} this. To see this, we first show how they generate the AMR graph in Figure~\ref{fig:previous_models}. Each time they update the graph, they first produce a new vertex (the vertex \textit{really} in this example), then they connect this new vertex to its parents (the edge \textit{degree} in this example). There is no guarantee that the new vertex is produced in the breadth-first order. For example, in Figure~\ref{fig:previous_mistake}, the models may incorrectly produce a second-layer child vertex \textit{school} before producing the last first-layer child vertex \textit{really}. While they cannot guarantee the breadth-first order, they achieve it in most steps in the parsing. This is because they encourage this order during training by always choosing the gold next-child in breadth-first order, shown also in Figure~\ref{fig:previous_mistake}.

In this paper, we design a new architecture such that the AMR graph is guaranteed to be generated in the breadth-first order. In each step, we introduce a \textbf{focused parent} vertex, and produce the next child for this parent vertex. We will not change this focused parent until we have produced all its children. After producing all its children, we change the focused parent to be the next in breadth-first order, guaranteeing that the graph is generated in the desired order.
%by the order of breadth-first in order to make sure that the generation of the whole graph is in this order.

\begin{figure*}[tbp]
    \centering
    \includegraphics[width=0.9\textwidth]{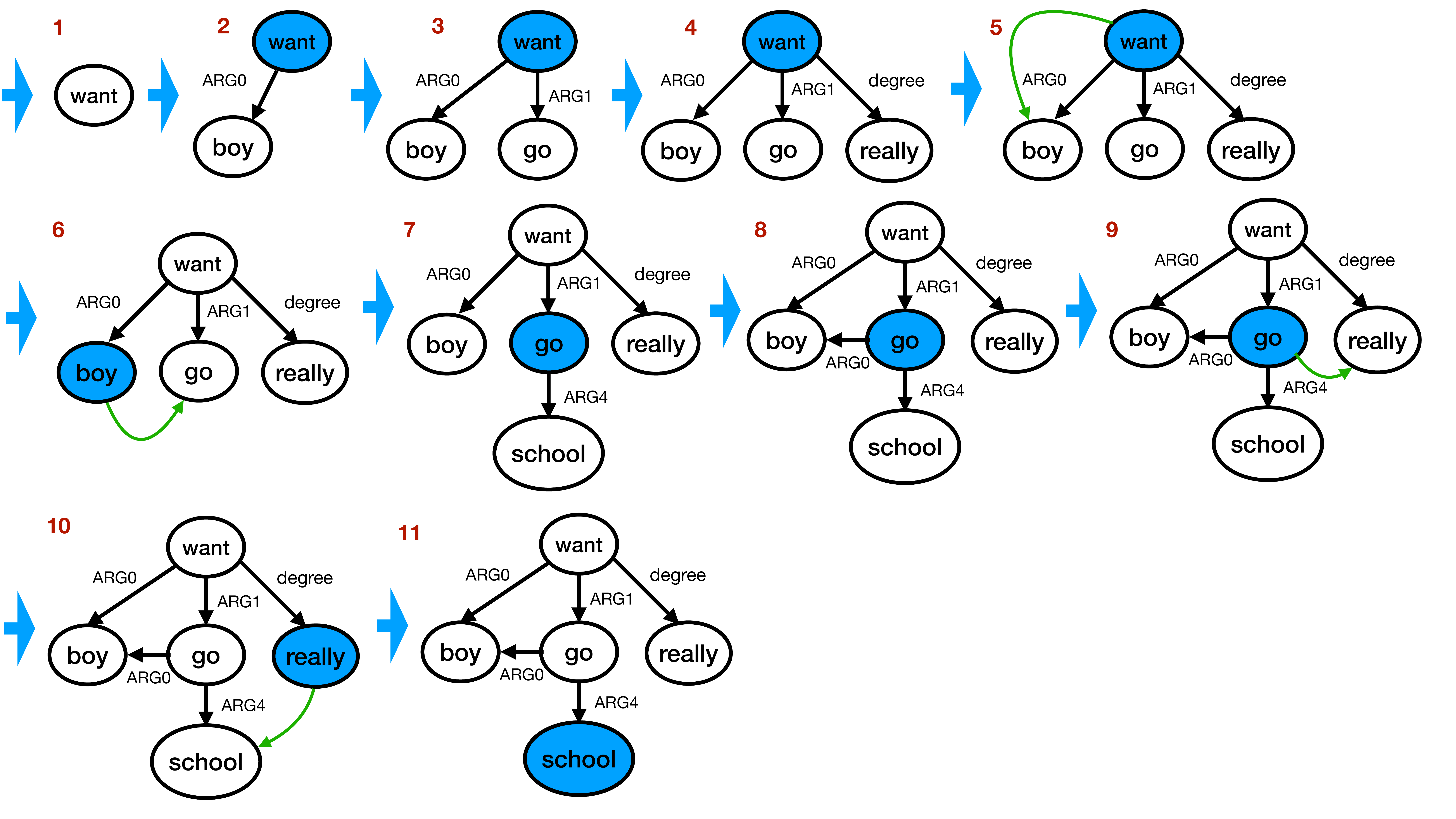}
    \caption{AMR Graph Producing Flow for Sentence: \textit{The boy really wants to go to school.} Here, the green arrows shows the shift of the focused parent vertex.}
    \label{fig:producing_flow}
\end{figure*}

For example, in Figure~\ref{fig:producing_flow}, we show the entire flow of producing the AMR graph for the sentence \textit{The boy really wants to go to school.} We first produce the root vertex \textit{want} (step 1), and label this root vertex as the \textbf{focused parent} vertex. Then we produce the three child vertices of the root vertex: \textit{boy}, \textit{go}, and \textit{really} (steps 2, 3, 4). In step 5, we find that the root vertex \textit{want} has no remaining children, so we shift our focus parent from \textit{want} to its first child \textit{boy}. In step 6, we find that \textit{boy} has no child, so we shift the focused parent from \textit{boy} to its sibling \textit{go}. In steps 7 and 8, we produce the two children of \textit{go}. One child is a new vertex \textit{school}, and the other is an existing vertex \textit{boy}. After that, we shift the focused parent to the last child of root vertex \textit{want} in step 9, but find no child of it in step 10. At this point, we have finished dealing with all the first-layer children. In step 11, we find there is no child of the only second-layer child \textit{school}. Therefore, we finish the whole parsing.

The contributions of our works are summarized as follows:
\begin{itemize}
    \item We propose a new AMR parsing architecture that strictly follows the breadth-first order to generate the AMR graph. In each step, we introduce a \textbf{focused parent} vertex to guide the construction.
    \item We improve the encoding part for the sentence and the partial graph by using a new \textbf{BERT Based Embedding Layer} (Section~\ref{sec:sentemb}) and adding edge information (Section~\ref{sec:graph_encoding}). We demonstrate their effectiveness in Section~\ref{sec:ablation}.
    \item Our model achieves better performance than \citeauthor{cai-lam-2020-amr} on both the AMR 1.0 dataset and the AMR 2.0 dataset.
\end{itemize}

\section{Some Concepts in Sentence and Graph} \label{sec:notation}
\begin{figure*}[tbp]
    \centering
    \subfloat[AMR from dataset]{
%        \centering
        \includegraphics[width=0.25\textwidth]{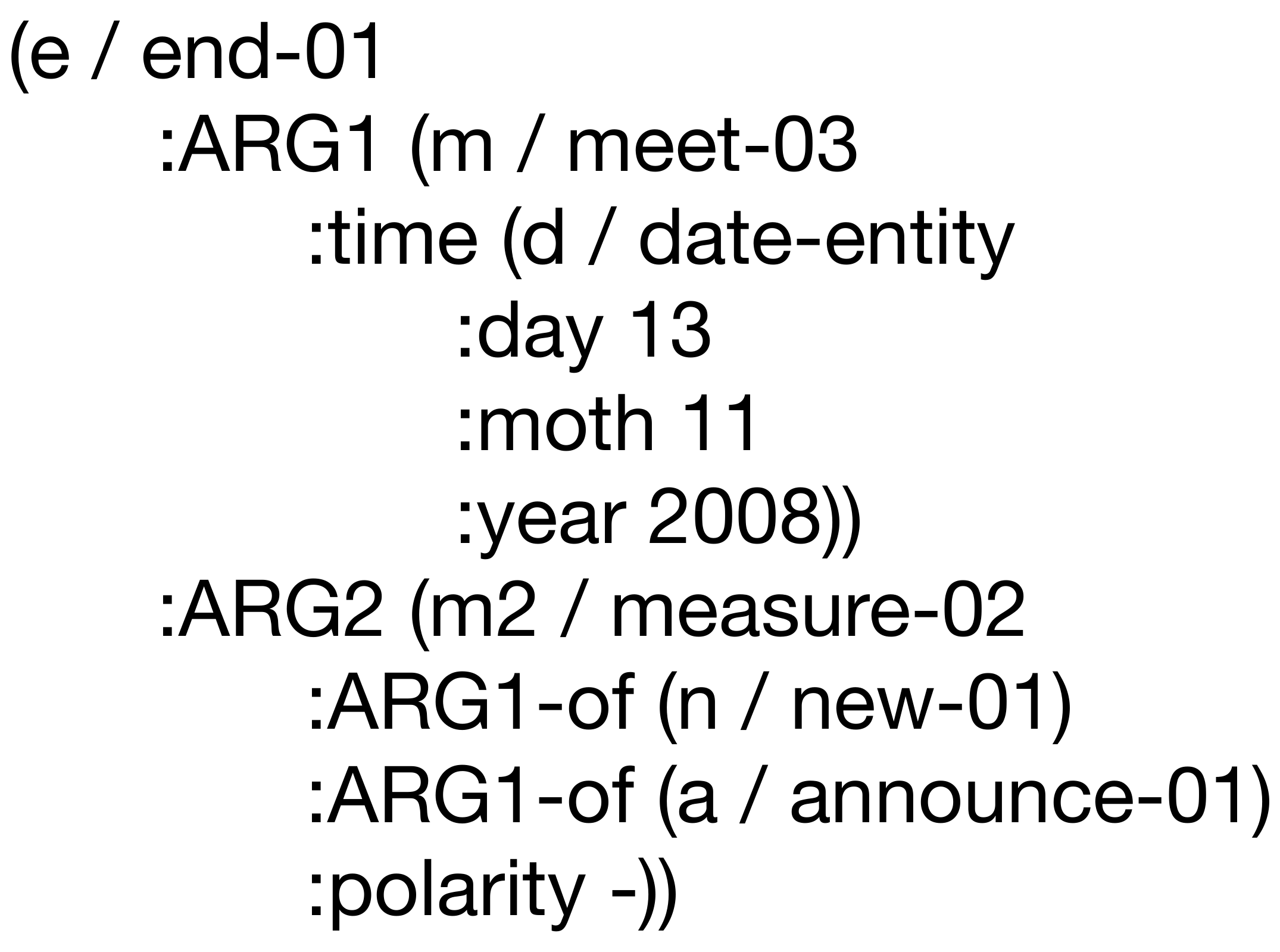}
%        \caption{AMR from dataset}
        \label{fig:amr-dataset}}
%    \end{subfloat}
%    \hfill
    \subfloat[AMR graph before processed]{
%        \centering
        \includegraphics[width=0.35\textwidth]{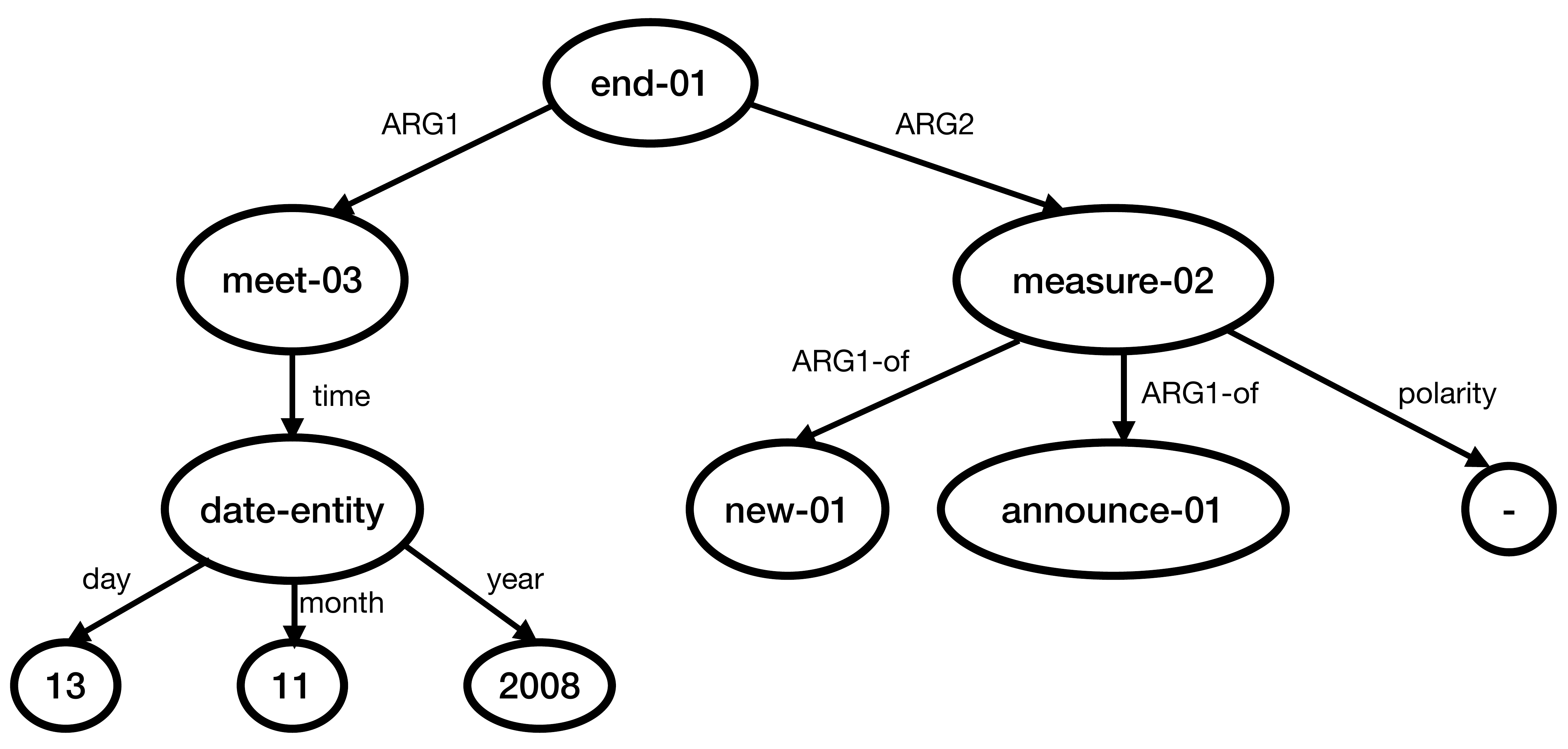}
%        \caption{AMR graph before processed}
        \label{fig:amr-before-processed}}
%    \end{subfloat}
%    \hfill
    \subfloat[AMR graph after processed]{
%        \centering
        \includegraphics[width=0.35\textwidth]{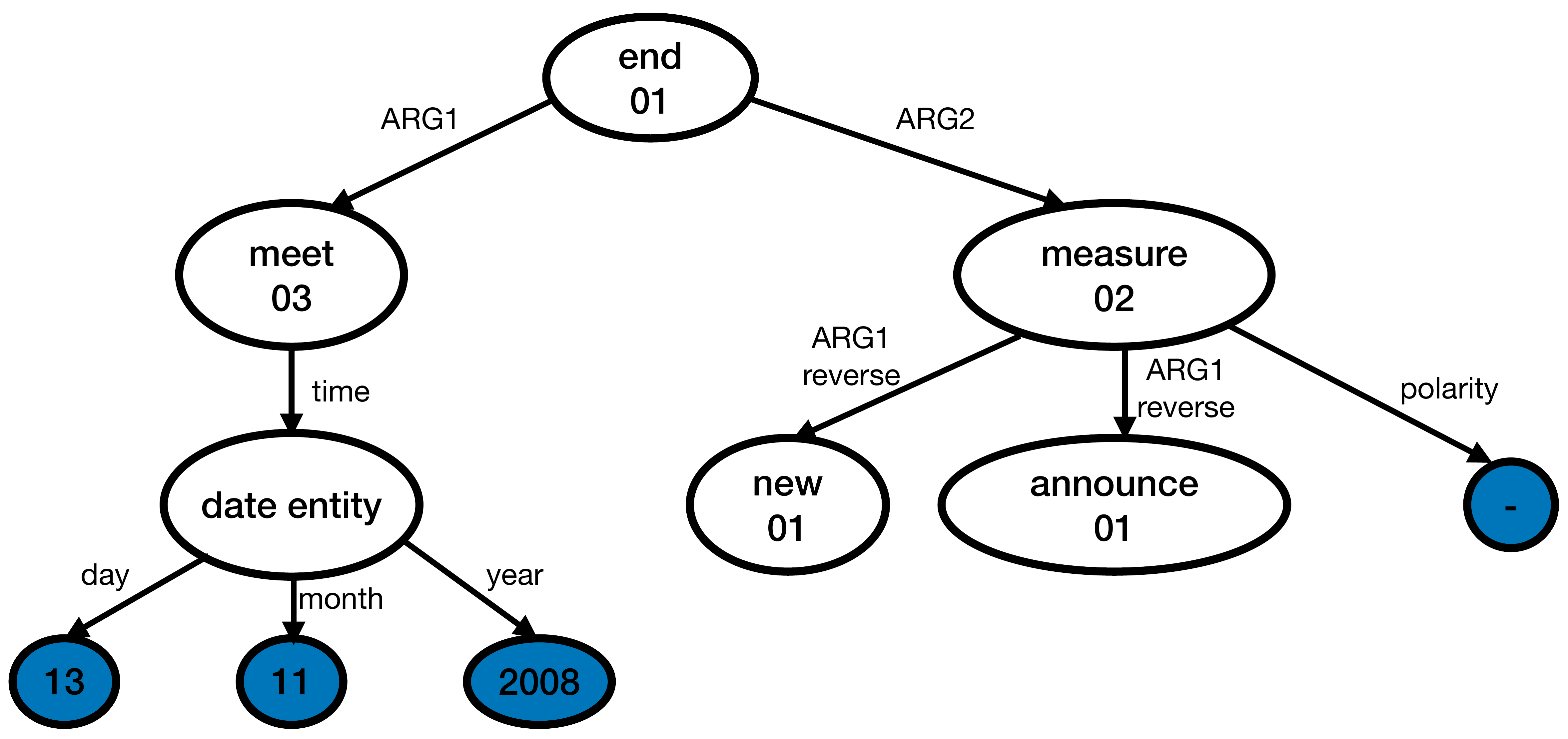}
%        \caption{AMR graph after processed}
        \label{fig:amr-after-processed}}
    \caption{AMR graph preprocess for the sentence: \textit{The 081113 meeting ended without any new measures to announce.} Notice that we show the \textit{type} part by the color. Blue vertices mean attribute vertices, and white vertices mean instance vertices}
    \label{fig:amr-preprocess}
\end{figure*}

In this section, we give some explanations about the vertices and edges in the graph under our definition. We divide every vertex and edge into several parts and will process or predict these parts separately in our model.  In Figure~\ref{fig:amr-dataset}, we show an AMR from our dataset and translate it into a graph in Figure~\ref{fig:amr-before-processed} before these processing steps. We show the graph after these processing steps in Figure~\ref{fig:amr-after-processed}.

\subsection{Vertex} \label{sec:vertex}
We divide every vertex into three parts: type, sense and content.

\paragraph{Type}
There are two types for each vertex, and we call them \textbf{instance} and \textbf{attribute}. We can see this in Figure~\ref{fig:amr-dataset}. In that AMR, some vertices are preceded by some variables, like \textit{e / end-01} (variable \textit{e}), \textit{d / date-entity} (variable \textit{d}), and \textit{n / new-01} (variable \textit{n}). These vertices are of type \textbf{instance}. The others do not have a preceding variable, like \textit{13}, \textit{2008}. These vertices are of type \textbf{attribute}. 

Intuitively, an \textbf{instance} is produced according to the meaning of a word or a phrase from the sentence, usually not a direct copy of some word (but sometimes is a copy from the \emph{lemma} of this word). Some special instances do not occur in the sentence, like \textit{date-entity} in Figure~\ref{fig:amr-dataset}. 

An \textbf{attribute} represents some property of its parent. For example, a \textit{name} instance vertex usually has attributes each representing one word of the name. In the example of Figure~\ref{fig:amr-dataset}, 
\textit{13}, \textit{11}, and \textit{2008} are the exact day, month, and year property of their parent, \textit{date-entity}. If an attribute occurs in the sentence, it is usually the same as some word in the sentence. But there are some attributes that are not the copy of words, like the negative property \textit{`-'} in this example.

\paragraph{Content and sense}
Some \textbf{instance}s end with some number called \textbf{sense}. The same word with different sense has different meanings. We separate an instance into \textbf{content} and \textbf{sense} (if present), by the dash. In Figure~\ref{fig:amr-dataset}, for example, \textit{end-01} is separated into \textit{end} and \textit{01}. 

Furthermore, in an \textbf{instance}, if its \textbf{content} is composed of more than one word, these words are usually connected with a hyphen, like \textit{date-entity}. We separate an instance by hyphens into several words.

\subsection{Edge} \label{sec:edge}
We divide the edge label into two parts: \textbf{label} and \textbf{if-reverse}.

Some edges in the AMR graph end with \textit{-of}. They represent the reverse relation. For example, in Figure~\ref{fig:amr-dataset}, \textit{measure-02} has a child \textit{new-01} with edge \textit{ARG1-of}. The sequential reading is that the \textit{ARG1-of} branch of \textit{measure-02} is \textit{new-01}, but the graph interpretation is that the \textit{ARG1} branch of \textit{new-01} is \textit{measure-02}. 

If an edge ends with \textit{-of}, we say this label's \textbf{if-reverse} part is true, otherwise it is false. The left part in this edge is called its \textbf{label}.

\section{Model Architecture} \label{sec:model}
In this section, we describe the details of our model. Given the source sentence, the  partial graph and the focused parent, the model produces the next child node of this focused parent (step 3 in Figure~\ref{fig:producing_flow}) or change the focused parent (step 5 in Figure~\ref{fig:producing_flow}).

Our model consists of four parts: \textbf{Sentence Encoder}, \textbf{Partial Graph Encoder}, \textbf{Interactive Module} and \textbf{Prediction Module}. The \textbf{Sentence Encoder} and \textbf{Partial Graph Encoder} separately encode the source sentence and the partial graph into some hidden vectors. The \textbf{Interactive Module} exchanges the information from the sentence and the partial graph with each other. The \textbf{Prediction Module} uses the exchanged information to make necessary predictions to build the next child or change the focused parent.

For the following discussion, let us assume that the source sentence is denoted by $w_1, w_2, \ldots, w_m$, where $w_i$ ($1 \leq i \leq m$) denotes the $i$th token in the sentence. The partial graph is denoted by $v_1, v_2, \ldots, v_t$, where $v_j$ ($1\leq j\leq t$) denotes the $j$th node in the partial graph. If $v_i$ has $v_j$ as its child, we use $e_{i,j}$ to denote the edge from $v_i$ to $v_j$, and also use $e_{j,i}$ to denote the reverse edge of $e_{i,j}$ (add or remove the suffix \textit{-of}). We use $\mathscr{E}_i$ to denote the set of edges that point to all the children and parents (in-neighbours and out-neighbours) of $v_i$.

\subsection{Sentence Encoder} \label{sec:sentencoder}
We encode the sentence $w_1, w_2, \ldots, w_m$ into hidden vectors $\bm{h}_0, \bm{h}_1, \bm{h}_2, \ldots, \bm{h}_m$ in this part. Note that there is one more vector $\bm{h}_0$, which is used to summarize the sentence and for the prediction module. To perform the encoding, we first compute the embedding vector $\bm{e}_i$ for each token $w_i$ in the sentence. 
%These are, lemma vector $\bm{l}_i$, token vector $\bm{t}_i$, part-of-speech vector $\bm{p}_i$, named entity tagging vector $\bm{n}_i$ and BERT vector $\bm{b}_i$. We average them as the total embedding vector $\bm{e}_i$. 
Then we use $\{\bm{e}_i\}_{i=1}^m$ to compute the encoding vectors $\{\bm{h}_i\}_{i=0}^m$.

\subsubsection{Embedding Vectors for the Sentence} \label{sec:sentemb}
We use a framework similar to \citet{cai-lam-2020-amr} for this part. In that paper, they computed five vectors for every token in the sentence, corresponding to the lemma $\bm{l}_i$, token itself $\bm{t}_i$, part-of-speech $\bm{p}_i$, named entity tag $\bm{n}_i$ and BERT feature $\bm{b}_i$, then combined these five vectors as the final embedding vector $\bm{e}_i$.

They run BERT for the input sentence and used the vectors of the top layer to construct the BERT feature $\bm{b}$. Since BERT is computed on sub-tokens instead of tokens of the sentence, they first computed BERT encoding vectors on sub-tokens of the sentence, then averaged the vectors of the sub-tokens within each token as the final vector of BERT feature. For the remaining four embedding vectors $\bm{l}, \bm{t}, \bm{p}$ and $\bm{n}$, they used four learned embedding layers to compute them.

% For the first four, they used four learnt embedding layers to compute them. For the BERT feature, they wanted to leverage BERT to help perform the sentence embedding. Since BERT is computed on sub-tokens instead of tokens of the sentence, they first computed BERT encoding vectors on sub-tokens of the sentence, then averaged the vectors of the sub-tokens within each token as the final vector of BERT feature.

Here, we use the same BERT feature $\bm{b}$ to summarize the information of the whole sentence. We also similarly use two learned embedding layers for part-of-speech $\bm{p}$ and name entity tag $\bm{n}$. But for the token $\bm{t}$ and lemma $\bm{n}$, we design an improved embedding layer called \textbf{BERT Based Embedding Layer}.

\paragraph{BERT Based Embedding Layer}
The reason to use a different method for the token and lemma is that they are words, not just a label like part-of-speech and name entity tag. The naive embedding layer in \citet{cai-lam-2020-amr} used randomly initialized learned vectors for every word (token and lemma). However, we want to leverage the powerful pretrained model --- BERT --- to encode the word, even though BERT is originally designed to encode the sentence, not a single word.
Therefore, we construct the final embedding vector using the sum of the BERT vector and a learned refinement embedding vector. The refinement embedding vector is used to adjust BERT to fit our specific AMR parsing task. Specifically, suppose we want to find the embedding vector for word $w$, and the BERT sub-tokenization for $w$ is $s_1, s_2, \ldots, s_u$. We use BERT to compute the encoding vector $\bm{s}_1, \bm{s}_2, \ldots, \bm{s}_u$ for each sub-token, then we compute the final BERT based embedding vector as follows:
\begin{eqnarray}
    % \begin{aligned}
        \bm{w}^{\textrm{BERT}} &=& \sum_{i=1}^u \bm{s}_i / u, \label{eq:bert}\\
        \bm{w}^{\textrm{Final}} &=& W^T\left(a\left(W^B\bm{w}^{\textrm{BERT}} + \textrm{Emb}\left(w\right)\right)\right), \nonumber
    % \end{aligned}
\end{eqnarray}
where $\bm{w}^{\textrm{BERT}} \in \mathbb{R}^b, \bm{w}^{\textrm{Final}} \in \mathbb{R}^g, W^B \in \mathbb{R}^{e\times b}, W^T\in \mathbb{R}^{g\times e}$, $a(\cdot)$ is the activation function and $\textrm{Emb}(\cdot)$ is the learnt refinement embedding layer. Here, $b$ is the length for BERT encoding vector, $e$ is the size for refinement embedding vector, $g$ is graph size, which will be the same as the length of graph encoding vectors and $u$ is the number of sub-tokens in word $w$. Here, we let $e$ be smaller than $b$ since we think the main part of embedding is still from BERT.

\textbf{BERT Based Embedding Layer} is used to compute the word embeddings, like token embedding $\bm{t}$ and lemma embedding $\bm{l}$. It is different from the BERT feature embedding vector $\bm{b}$ (remember that we have five different embedding parts: token, lemma, part-of-speech, name entity tagging and BERT feature). For the BERT feature $\bm{b}$, we compute the vectors based the whole sentence information, but the \textbf{BERT Based Embedding Layer} only considers the word itself. Note that our model still uses the BERT feature $\bm{b}$ to summarize the information of the whole sentence.

\subsubsection{Sentence Encoding}
 We insert a special embedding $\bm{e}_0$ at the beginning of the sentence in order to summarize the sentence later. Now, the sequence embedding vectors become $\{\bm{e}_i\}_{i=0}^m$. We then employ $L_S$ Transformer Encoder Layers to encode the sentence. We denote the final sentence encoding vector by $\{\bm{h}_i\}_{i=0}^m$.

\subsection{Partial Graph Encoder}
Suppose now we have $t$ vertices in the partial graph, denoted by $v_1, v_2, \ldots, v_t$, and some edges between them denoted by $\{e_{i,j}\}$ (in the beginning of Section~\ref{sec:model} we mentioned that we have added the reverse edge for every edge). We will encode this partial graph into $t+1$ hidden vectors $\{\bm{g}_i\}_{i=0}^t$ in this part, where each hidden vector $\bm{g}_i$ denotes the vertex $v_i$ and the information about its descendants and ancestors, and we will also insert a special encoding vector $\bm{g}_0$ as in the sentence encoding.

To do this, we first compute vertex and edge embedding vectors $\bm{v}_i$ and $\bm{e}_{i,j}$ for each vertex $v_i$ and edge $e_{i,j}$. Then we use a Graph Recurrent Network to gradually collect information about the descendants' and ancestors' information.

\subsubsection{Vertex Embedding} \label{sec:vertex_embedding}
In Section~\ref{sec:vertex}, we mentioned that a vertex is composed of three parts: type, content, and sense. Here, we only use content information for graph encoding. 
%But we will still mention how to compute embedding vectors for type and sense since they will be used in \textbf{Predict Module}.

For illustration, we consider the instance vertex \textit{go-back-19}. As we discussed in Section~\ref{sec:notation}, the content of this node is \textit{go back}. We use \textbf{BERT Based Embedding Layer} to compute the content embedding $\bm{v}_i$. Here, even though there are two words in \textit{go back}, we are still able to use \textbf{BERT Based Embedding Layer} since the sub-token algorithm in BERT can also be used on phrases, and for the refinement part, we can treat the phrase as a whole unit and assign an embedding vector for this unit.
%We think the sense information is included in the BERT pretraining and the class information is not important for graph encoding. Like sentence encoder, we also add a special embedding $\bm{v}_0$ at the beginning of the graph node sequence.

%\paragraph{Type and Sense Embedding}
%We use two naive learnt embedding layers for type and sense embedding. We denote the type embedding vector by $\bm{p}_i$ and sense embedding vector by $\bm{s}_i$.

\subsubsection{Edge Embedding} \label{sec:edge_embedding}
For the edge $e_{i, j}$, as we introduced in Section~\ref{sec:edge}, it is composed of the label part and if-reverse part. We use two naive embedding layers to separately compute their embeddings $\bm{e}_{i,j}^{\textrm{label}}$ and $\bm{e}_{i,j}^{\textrm{reverse}}$, and compute their sum as the total edge embedding $\bm{e}_{i,j}$.

\subsubsection{Graph Encoding} \label{sec:graph_encoding}
Having the embeddings of all the vertices and edges, we are able to encode this partial graph. We construct the graph encoding repeatedly by gradually collecting the information about the descendants and ancestors for every vertex. Formally, we repeat the following recurrence for $L_G$ times:
\begin{equation*}
\begin{aligned}
    \bm{v}_i^{l,0} =& \textrm{LN}\left(\bm{v}_i^{l-1,1} + \sum_{j\in \mathscr{E}_i}W^l[\bm{v}_j^{l-1,1}; \bm{e}_{i,j}] / |\mathscr{E}_i| \right), \\
    \bm{v}_i^{l,1} =& \textrm{LN}\left(\bm{v}_i^{l,0} + \textrm{FFN}^{l}\left(\bm{v}_i^{l,0}\right)\right), 
\end{aligned}
\end{equation*}
where LN represents layer normalization layer and FFN denotes feed-forward layer, and $\bm{v}_i^{0,1} = \bm{v}_i$. We use the output of the last layer $\{\bm{v}_i^{L_G, 1}\}_{i=1}^t$ as the graph encoding result, and we denote them by $\{\bm{g}_i\}_{i=1}^t$. Then, as we did for the sentence encoding, we insert a special encoding vector $\bm{g}_0$.

Different from \citet{cai-lam-2020-amr}, we add edge information here. We will demonstrate its effectiveness in Section~\ref{sec:ablation}.

\subsection{Interactive Module}
After we obtain the sentence encoding vectors $\{\bm{h}_i\}_{i=0}^m$ and partial graph encoding vectors $\{\bm{g}_i\}_{i=0}^t$, we want to exchange these pieces of information and prepare a vector for the \textbf{Prediction Module}. We also include the information about the \textbf{focused parent} in this module. Suppose the focused parent is $v_{\textrm{focus}}$ and its encoding vector is $\bm{g}_{\textrm{focus}}$. We design an \textbf{interactive Transformer layer} to let them interact with each other. 

\paragraph{Interactive Transformer Layer} Specifically, in the $l$th interactive Transformer layer, we first add weighted sentence information to each graph vertex:
\begin{equation*}
\begin{aligned}
    \bm{g}_i^{l,0} &= \textrm{LN}\left(\bm{g}_i^{l-1,2} + \textrm{Attn}^{l,S,0}\left(\bm{g}_i^{l-1,2}, \bm{h}_{0:m}^{l-1,3}\right)\right), \\
    \bm{g}_i^{l,1} &= \textrm{LN}\left(\bm{g}_i^{l,0} + \textrm{Attn}^{l,S,1}\left(\bm{g}_i^{l,0}, \bm{g}_{0:t}^{l,0}\right)\right), \\
    \bm{g}_i^{l,2} &= \textrm{LN}\left(\bm{g}_i^{l,1} + \textrm{FFN}^{l,S}\left(\bm{g}_i^{l,1}\right)\right),
\end{aligned}
\end{equation*}
where $\bm{h}_i^{0,3} = \bm{h}_i$, and $\bm{g}_i^{0,2} = \bm{g}_i$. Then, we add weighted graph vertex information to each sentence node. We also add the \textbf{focused parent} information here:
\begin{equation*}
\begin{aligned}
    \bm{h}_i^{l,0} &= \textrm{LN}\left(\bm{h}_i^{l-1,3} + \textrm{Attn}^{l,G,0}\left(\bm{h}_i^{l-1,3}, \bm{g}_{0:t}^{l,2}\right)\right), \\
    \bm{h}_i^{l,1} &= \textrm{LN}\left(\bm{h}_i^{l,0} + \bm{g}_{\textrm{focus}}\right), \\
    \bm{h}_i^{l,2} &= \textrm{LN}\left(\bm{h}_i^{l,1} + \textrm{Attn}^{l,G,1}\left(\bm{h}_i^{l,1}, \bm{h}_{0:m}^{l,1}\right)\right), \\
    \bm{h}_i^{l,3} &= \textrm{LN}\left(\bm{h}_i^{l,2} + \textrm{FFN}^{l,G}\left(\bm{h}_i^{l,2}\right)\right). \\
\end{aligned}
\end{equation*}

~\\
\noindent We apply $L_I$ interactive Transformer layers. We use the first vector of the sentence in the top layer $\bm{h}_0^{L_I, 3}$ for prediction and we denote it by $\bm{h}^p$. Also, we will use $\{\bm{g}_i^{L_I, 2}\}_{i=1}^t$ in the \textbf{Prediction Module} as well, and we denote them by $\{\bm{g}_i^p\}_{i=1}^t$.

\subsection{Prediction Module} \label{sec:predict}
Finally, we come to the prediction part. First, we use $\bm{h}^p$ (obtained in \textbf{Interactive module}) to predict if there is another child for the focused parent. If there is no other child, we will change the focused parent. Otherwise, we will produce the vertex and edge of the next child.

We first predict the vertex, then predict the edge based on the predicted vertex. If the next child is a vertex produced previously, we will use $\bm{h}^p$ and the information of all produced vertices $\{\bm{g}_i^p\}_{i=1}^t$ (obtained in the \textbf{Interactive module}) to predict it. If the next vertex is a new instance or attribute, we will use $\bm{h}^p$ to predict the content and sense of the new vertex. Like \citet{cai-lam-2020-amr}, the prediction of the content can be either producing a new content from content vocabulary or copying of some word in sentence.

After predicting the vertex, we use it together with $\bm{h}^p$ to predict the new edge. Like the vertex, we separately predict the \textit{label} part and \textit{if-reverse} part of the edge.

The details of this Prediction Module are provided in Appendix~\ref{sec:detail_predict}. 

%If the child node is a previously-produced node, for example, node $i$, we will use the graph encoding vector $\bm{g}_i^f$ to represent it. If the child node is a new node, we will use the graph node embedding (add this part in section 4.2) $\bm{b}_{t+1}$ to represent it. Anyway, we use $\bm{n}^{predict}$ to denote the vector. Then, we have:
%\begin{equation*}
%    P^{edge} = \textrm{Softmax}\left(W^{edge}\left(\bm{h}_f, \bm{n}^{predict}\right) + \bm{b}^{edge}\right).
%\end{equation*}

\section{Training and Inference}
\subsection{Training}
We use the standard maximum likelihood method to train the model. The loss function for one data point (input sentence with gold graph) is the sum of the negative log-likelihood for every partial graph corresponding to the gold graph, where the negative log-likelihood for each partial graph is again the sum of all the components introduced in Section~\ref{sec:predict} (and Appendix~\ref{sec:detail_predict} for details). We use breadth-first order to break the gold graph into several incremental partial graphs. But for each parent, there is still more than one order to produce its children. Following \citet{cai-lam-2020-amr},  with 0.5 probability we use a deterministic order (sorted by the frequency of edge labels), and with 0.5 probability we use a uniformly random order. 

\subsection{Inference}
Since our model is auto-regressive (it produces a new vertex and edge based on previous partial graphs and in each partial graph predicts a new object based on previous predicted objects, see Section~\ref{sec:predict}), we use beam search to produce the whole graph.

\section{Experiments}
\subsection{Setup}
\paragraph{Dataset}
We evaluate our model on the AMR 1.0 (LDC2014T12) and  AMR 2.0 (LDC0217T10) datasets. AMR 1.0 contains 13051 sentences in total while AMR 2.0 has about 39000. Both datasets have already been split into training, development and testing parts. Since AMR 2.0 is larger, we use it as our main dataset.
\paragraph{Implementation Detail}
We use Stanford CoreNLP \citep{manning:2014} to obtain the token, lemma, part-of-speech and named-entity tagging for each word in the sentence (Section~\ref{sec:sentencoder}). We use pre-trained models in \citet{devlin2019bert} to compute BERT vectors.

We employ similar pre-processing and post-processing steps to \citet{cai-lam-2020-amr}. For the pre-processing step, we also break the vertex and the edge into several parts as described in Section~\ref{sec:notation}. For the post-processing, we also employ the reverse operations of pre-processing operations, combining the several components into a whole unit for the vertex and the edge. 
%One thing different from almost all the previous paper is that we remove all the \textit{:wiki} attributes in our dataset and our prediction. We think wikification has nothing to do with the model evaluation and we will explain this later (Section~\ref{sec:nowiki}).

During training, we fix the BERT parameters similar to \citet{cai-lam-2020-amr} due to GPU limitation (and we add the learned refinement term in BERT Based Embedding Layer for compensation). We use ADAM optimization with decayed learning rate similar to \citet{vaswani-nips17} to train the model. Training takes approximately 7 hours on four Nvidia GeForce GTX 1080 Ti.

We use development data to tune the hyper-parameters. The hyper-parameters we used in our best model are described in Appendix~\ref{sec:hyperparameter}. We use a beam size of 8 to produce the graph. \footnote{We will release code with the final version.}

\begin{table*}[tbp]
    \centering
    \setlength\arrayrulewidth{1pt}
    \scalebox{0.85}{
    \begin{tabular} {|c||c||c||c|c|c|c|c|c|c|c|}
        \hline
        model & G.R. & Smatch & Unlabeled & NO WSD & Concept & SRL & Reent. & Neg. & NER & wiki\\
        \hline\hline
        % \citet{van2017neural} & $\times$  & 71.0 & 74 & 72 & 82 & 66 & 52 & 62 & 79 & 65\\
        % \hline
        % \citet{cai2019core} & $\times$  & 73.2 & 77.0 & 74.2 & 84.4 & 66.7 & 55.3 & 62.9 & 82.0 & 73.2 \\
        % \hline
        % \citet{cai-lam-2020-amr} & $\times$  & 78.7 & 81.5 & 79.2 & 88.1 & 74.5 & 63.8 & 66.1 & 87.1 & 81.3 \\
        % \hline\hline
        % Ours & $\times$ & 80.0 & \textbf{83.1} & \textbf{80.8} & 87.4 & \textbf{75.6} & \textbf{67.5} & 62.5 & \textbf{87.5} & 79.7 \\
        % \hline\hline
        \citet{groschwitz-acl18} & $\checkmark$  & 71.0 & 74 & 72 & 84 & 64 & 49 & 57 & 78 & 71\\
        \hline
        \citet{lyu-acl18} & $\checkmark$  & 74.4 & 77.1 & 75.5 & 85.9 & 69.8 & 52.3 & 58.4 & 86.0 & 75.7 \\
        \hline
        \citet{lindemann-etal-2019-compositional} & $\checkmark$  & 75.3 & - & - & - & - & - & - & - & - \\
        \hline
        \citet{naseem2019rewarding} & $\checkmark$  & 75.5 & 80 & 76  & 86  & 72 & 56 & 67 & 83 & 80 \\
        \hline
        \citet{zhang-etal-2019-amr} & $\checkmark$ & 76.3 & 79.0 & 76.8 & 84.8 & 69.7 & 60.0 & 75.2 & 77.9 & 85.8 \\
        \hline
        \citet{zhang-etal-2019-broad} & $\checkmark$  & 77.0 & 80 & 78 & 86 & 71 & 61 & 77 & 79 & 86 \\
        \hline
        \citet{zhou-etal-2020-amr} & $\checkmark$ & 77.5 & 80.4 & 78.2 & 85.9 & 71.0 & 61.1 & 76.1 & 78.8 & 86.5 \\
        \hline
        \citet{cai-lam-2020-amr} & $\checkmark$ & 80.2 & 82.8 & 80.8 & \textbf{88.1} & 74.2 & 64.6 & \textbf{78.9} & \textbf{81.1} & 86.3 \\
        \hline
        Ours & $\checkmark$ & \textbf{81.1} & \textbf{84.0} & \textbf{81.9} & 87.3 & \textbf{74.9} & \textbf{67.7} & 76.7 & 80.4 & \textbf{86.9} \\
        \hline
    \end{tabular}
    }
    \caption{Overall Smatch scores and fine-grained scores of different parsers on AMR 2.0 dataset. Some results of previous models come from \citet{cai-lam-2020-amr}. Here, G.R. means graph re-categorization. All the models in this table include the re-categorization process.}
    \label{tab:wiki_result}
\end{table*}

\subsection{Results}
\paragraph{Main Results}
We show our main results in Table~\ref{tab:wiki_result}. This includes the overall Smatch score and fine-grained scores \citep{DamonteCS17} of our model and previous models on the AMR 2.0 dataset. We can see that our model beats \citeauthor{cai-lam-2020-amr} by 0.9\% and is better for the most of the fine-grained scores. %for the first time over 80\%. Even compared with model that have re-categorization pre-processing, our model beats almost every previous model, except the model in \citet{cai-lam-2020-amr}. In fact, in Table~\ref{tab:nowiki_result}, when we remove the \textit{wiki} attribute in all gold and predicted AMR graphs, our model obtains the highest Nowiki-Smatch score. For most fine-grained scores, our model creates SOTA performance.

\paragraph{Results for Models without Re-categorization}
Most of the AMR parsing models include a complex pre-processing step called \textbf{re-categorization}. With re-categorization, speciﬁc sub-graphs of a AMR graph (usually corresponding to special entities, like named entities, date entities, etc.) are treated as a unit and assigned to a single vertex with a new content. 
%In Figure~\ref{fig:recategorization}, we show the AMR graph of sentence \textit{By the 10th Century AD, 60\% of Britain's territory belonged to the Church} before and after re-categorization process. In Figure~\ref{fig:before_recategorization}, we use four colors to denote the four sub-graphs which need re-categorization and use the same color to label the corresponding word spans in the sentence. In Figure~\ref{fig:after_recategorization}, we show the four new single vertices corresponding the sub-graphs in Figure~\ref{fig:before_recategorization} with the same color, as well as the four new word spans in the sentence.
The re-categorization process is composed of several hand-crafted rules, requiring exhaustive screening and expert-level manual efforts. This is a feasible idea for the current dataset since the dataset is small (about 40000 sentences in AMR 2.0 dataset). %only having limited ways of re-categorization and recovery from re-categorization. 
However, when the dataset becomes larger in the future, using machine learning to automatically learn a model will be a better idea than hand-crafted rules.

Therefore, we want to see the performance of our model without re-categorization and we show the result in Table~\ref{tab:wiki_result_no_recategorization}. Our model beats \citeauthor{cai-lam-2020-amr} by 1.3\% and is above 80\% for the first time. Also, our model is better for most of the fine-grained scores.

\begin{table*}[tbp]
    \centering
    \setlength\arrayrulewidth{1pt}
    \scalebox{0.85}{
    \begin{tabular} {|c||c||c||c|c|c|c|c|c|c|c|}
        \hline
        model & G.R. & Smatch & Unlabeled & NO WSD & Concept & SRL & Reent. & Neg. & NER & wiki\\
        \hline\hline
        \citet{van2017neural} & $\times$  & 71.0 & 74 & 72 & 82 & 66 & 52 & 62 & 79 & 65\\
        \hline
        \citet{cai2019core} & $\times$  & 73.2 & 77.0 & 74.2 & 84.4 & 66.7 & 55.3 & 62.9 & 82.0 & 73.2 \\
        \hline
        \citet{cai-lam-2020-amr} & $\times$  & 78.7 & 81.5 & 79.2 & \textbf{88.1} & 74.5 & 63.8 & \textbf{66.1} & 87.1 & \textbf{81.3} \\
        \hline\hline
        Ours & $\times$ & \textbf{80.0} & \textbf{83.1} & \textbf{80.8} & 87.4 & \textbf{75.6} & \textbf{67.5} & 62.5 & \textbf{87.5} & 79.7 \\
        \hline\hline
    \end{tabular}
    }
    \caption{Overall Smatch scores and fine-grained scores of different parsers on AMR 2.0 dataset. Some results of previous models come from \citet{cai-lam-2020-amr}. Here, G.R. means graph re-categorization. All the models in this table do not include the re-categorization process.}
    \label{tab:wiki_result_no_recategorization}
\end{table*}

\paragraph{Results for AMR 1.0 Dataset}
Next we evaluate our model on the AMR 1.0 dataset. We show the results in Table~\ref{tab:amr1_result_with_reca} for models with the re-categorization pre-processing step and show the results in Table~\ref{tab:amr1_result_without_reca} for models without the re-categorization pre-processing step. We can see that our model beats every previous model for both situations.

\begin{table}[tbp]
    \centering
    \setlength\arrayrulewidth{1pt}
    \scalebox{1.0}{
    \begin{tabular} {|c||c|c|}
        \hline
        model & G.R. & Smatch \\
        \hline
        % \citet{flanigan-acl14} & $\times$ & 66.0 \\
        % \hline
        % \citet{pust-etal-2015-parsing} & $\times$ & 67.1 \\
        % \hline
        % \citet{cai-lam-2020-amr} & $\times$ & 74.0 \\
        % \hline
        % ours & $\times$ & \textbf{74.6} \\
        % \hline
        \citet{wang-xue-2017-getting} & $\checkmark$ & 68.1 \\
        \hline
        \citet{guo-lu-2018-better} & $\checkmark$ & 68.3 \\
        \hline
        \citet{zhang-etal-2019-amr} & $\checkmark$ & 70.2 \\
        \hline
        \citet{zhang-etal-2019-broad} & $\checkmark$ & 71.3 \\
        \hline
        \citet{cai-lam-2020-amr} & $\checkmark$ & 75.4 \\
        \hline
        Ours & $\checkmark$ & \textbf{75.8} \\
        \hline\hline
    \end{tabular}
    }
    \caption{Smatch scores of different models with the re-categarization pre-processing step on AMR 1.0 dataset. Some results of previous models come from \citet{cai-lam-2020-amr}.}
    \label{tab:amr1_result_with_reca}
\end{table}

\begin{table}[tbp]
    \centering
    \setlength\arrayrulewidth{1pt}
    \scalebox{1.0}{
    \begin{tabular} {|c||c|c|}
        \hline
        model & G.R. & Smatch \\
        \hline
        \citet{flanigan-acl14} & $\times$ & 66.0 \\
        \hline
        \citet{pust-etal-2015-parsing} & $\times$ & 67.1 \\
        \hline
        \citet{cai-lam-2020-amr} & $\times$ & 74.0 \\
        \hline
        Ours & $\times$ & \textbf{74.6} \\
        % \hline
        % \citet{wang-xue-2017-getting} & $\checkmark$ & 68.1 \\
        % \hline
        % \citet{guo-lu-2018-better} & $\checkmark$ & 68.3 \\
        % \hline
        % \citet{zhang-etal-2019-amr} & $\checkmark$ & 70.2 \\
        % \hline
        % \citet{zhang-etal-2019-broad} & $\checkmark$ & 71.3 \\
        % \hline
        % \citet{cai-lam-2020-amr} & $\checkmark$ & 75.4 \\
        % \hline
        \hline\hline
    \end{tabular}
    }
    \caption{Smatch scores of different models without the re-categarization pre-processing step on AMR 1.0 dataset. Some results of previous models come from \citet{cai-lam-2020-amr}.}
    \label{tab:amr1_result_without_reca}
\end{table}

% \paragraph{Analysis of re-categorization process}
% Comparing the results of two datasets, we can see the limitation of re-categorization process. In the small AMR 1.0 dataset (only with about 13000 data), the best model with re-categorization process \citep{cai-lam-2020-amr} is better than our model. However, when testing in the larger AMR 2.0 dataset (with about 40000 data), our model is in turn better. This shows, the ability of re-categorization process fades away along with the increase of dataset size. This suggests us to switch our focus to the model without re-categorization process in the future.

\begin{table}[tbp]
    \centering
    \setlength\arrayrulewidth{1pt}
    \scalebox{1.0}{
    \begin{tabular} {|c|c|}
        \hline
        model & Smatch \\
        \hline
        with edge & 81.1 \\
        \hline
        without edge & 80.8\\
        \hline
    \end{tabular}
    }
    \caption{The effect of edge information.}
    \label{tab:edge_result}
\end{table}

\begin{table}[tbp]
    \centering
    \setlength\arrayrulewidth{1pt}
    \scalebox{1.0}{
    \begin{tabular} {|c|c|c|}
        \hline
        model & Smatch \\
        \hline
        our model & 81.1 \\
        \hline
        remove BERT & 80.7\\
        \hline
    \end{tabular}
    }
    \caption{Effect of BERT for \emph{BERT Based Embedding Layer}.}
    \label{tab:embedding_result}
\end{table}

\subsection{Ablation Study} \label{sec:ablation}
\paragraph{Effect of Edge Information}
In the previous SOTA model \citep{cai-lam-2020-amr}, they did not add edge information in Graph Encoder part. They claimed that the edge information had little influence on the performance of their model. Therefore, we want to see if edge information is important in our model. We evaluate it on AMR 2.0 dataset. From Table~\ref{tab:edge_result}, we see that the Smatch score of our model with edge information is higher than the score without edge information. This demonstrates that the edge information is helpful in our model.

\paragraph{Effect of BERT Based Embedding Layer} In our model, we introduce a new embedding layer named \textbf{BERT Based Embedding Layer} to better encode the tokens, lemmas (Section~\ref{sec:sentemb}) and vertices (Section~\ref{sec:vertex_embedding}). However, in the previous SOTA model \citep{cai-lam-2020-amr}, they only use learnt embedding layers for them, which means they did not include a BERT component (Equation~(\ref{eq:bert})) for those words (but they still run the BERT for the whole sentence to construct the BERT feature $\bm{b}$ mentioned in Section~\ref{sec:sentemb}). So we want to see its effect in our model. In Table~\ref{tab:embedding_result}, we can see that when removing the BERT part, the Smatch score decreases by 0.4\%, which demonstrates that BERT plays an important role here.

\section{Conclusion}
% \begin{itemize}
%     \item We propose a new AMR parsing architecture that strictly follow the breadth-first order to generate the AMR graph. In each step, we introduce a \textbf{focused parent} vertex to guide the construction.
%     \item We improve the encoding part for the partial graph by using a new \textbf{BERT Based Embedding Layer} (Section~\ref{sec:sentemb}) and adding edge information (Section~\ref{sec:graph_encoding}). We demonstrate their effectiveness in Section~\ref{sec:ablation}.
%     \item Our model achieves a new SOTA performance on both AMR 1.0 dataset and AMR 2.0 dataset.
% \end{itemize}
In this paper, we propose a new AMR parsing architecture that strictly follows the breadth-first order to generate the AMR graph. With the help of the focused parent information, a better word embedding layer named BERT Based Embedding Layer and the edge information in graph encoding, our model is able to construct the AMR more accurately. Our model achieves better performance on the both AMR 1.0 and AMR 2.0 datasets.

\paragraph{Acknowledgments} Research supported by NSF awards IIS-1813823 and CCF-1934962.

\bibliography{all}
\bibliographystyle{acl_natbib}

\newpage

\appendix

\section{Details of Prediction Module} \label{sec:detail_predict}

As we discussed in Section~\ref{sec:predict}, we first predict the vertex, then predict the edge based on the predicted vertex (if it exists).

\subsection{Predict Vertex}
There are four situations about the next vertex (we give each situation a label for classification):
\begin{enumerate}
    \item there is no other child (label 0);
    \item the next vertex is a node produced previously (label 1);
    \item the next vertex is a new instance (label 2);
    \item the next vertex is a new attribute (label 3).
\end{enumerate}
We must predict the situation before any further prediction. We use $\bm{o}$ to denote the status and use $\bm{h}^p$ to predict the status $\bm{o}$:
\begin{equation*}
    p(\bm{o}) = \textrm{Softmax}\left(W^o\bm{h}^p + \bm{b}^o\right).
\end{equation*}
Now we make further prediction based on $\bm{o}$.

\paragraph{No Other Child ($\bm{o} = 0$)} This means there are no other children, so we will change the focused parent, like the step 5 in Figure~\ref{fig:producing_flow}.

\paragraph{Previously Produced ($\bm{o} = 1$)} This means the next child is some vertex produced previously. We compute the probability vector using an Attention layer:
\begin{equation*}
    p_{1:t}^{r} = \textrm{Softmax}\left(\left(W^{r, 0}\bm{h}^p\right)^\top W^{r, 1}\bm{g}_{1:t}^p\right).
\end{equation*}
We use $v^r$ to denote the index of the produced vertex that the next child equals, then we have $p\left(v^r\right) = p_{v^r}^r$.

\paragraph{New Instance ($\bm{o} = 2$)} This means the next vertex is of type \textit{instance}. Then we must predict its content and sense. But before that, we use a simple linear layer to transfer $\bm{h}^p$ into instance mode $\bm{h}^{\textrm{conc}}$:
\begin{equation*}
    \bm{h}^{\textrm{conc}} = \bm{h}^p + a\left(W^{\textrm{conc}}\bm{h}^p\right),
\end{equation*}
We divide the prediction of content into two parts. In the first part, we think the content is a new word or a new phrase, not directly copied from any word in the sentence, such as \textit{date-entity}. In the second part, we think the content comes directly from the sentence. We use a soft linear layer to predict this:
\begin{equation*}
    [p_0, p_1] = \textrm{Softmax}\left(W^s\bm{h}^{\textrm{conc}} + \bm{b}^s\right).
\end{equation*}
Now, suppose we denote the predicted content by $c$. In the first part, we use linear and softmax layer to predict it from content vocabulary:
\begin{equation*}
    p^{\textrm{new}}(c) = \textrm{Softmax}(W^{\textrm{new}}\bm{h}^{\textrm{conc}} + \bm{b}^{\textrm{new}}).
\end{equation*}
In the second part, we think the content comes from some word (more precisely, the lemma of the word) of the sentence, so we use Attention layer to predict the probability vector: 
\begin{equation*}
    p_{1:m}^\textrm{lemma} = \textrm{Softmax}\left(\left(\bm{h}^{\textrm{conc}}\right)^\top\bm{l}_{1:m}\right),
\end{equation*}
where $\bm{l}_{1:m}$ are the embedding vectors of lemmas which are introduced in Section~\ref{sec:sentencoder}.
The final probability of the predicted content $c$ is:
\begin{equation*}
    p(c) = p_0p^{\textrm{newc}}(c) + p_1\sum_{i:l_i = c}p_i^{\textrm{lemma}}.
\end{equation*}

Once we predict the content, we use \textbf{BERT Based Embedding Layer} to obtain the embedding vector $\bm{c}$. Then we are able to predict the sense of this vertex. We use $s$ to denote the sense and compute the probability as follows:
\begin{equation*}
    p(s) = \textrm{Softmax}\left(W^{\textrm{sense}}\left[\bm{h}^{\textrm{conc}}; \bm{c}\right] + \bm{b}^{\textrm{sense}}\right).
\end{equation*}

\paragraph{New Attribute ($\bm{o}=3$)} This means the next vertex is of type \textit{attribute}. Notice that an attribute does not have a sense, so we only have to predict its content. Similar to \textit{instance}, we first transfer $\bm{h}^p$ into attribute mode $\bm{h}^{\textrm{attr}}$:
\begin{equation*}
    \bm{h}^{\textrm{attr}} = \bm{h}^p + a\left(W^{\textrm{attr}}\bm{h}^p\right).
\end{equation*}
Then we divide the prediction of attribute into two parts. One is a new word and the other is directly copying from some word in the sentence. Similar to \textit{instance}, we use the same soft linear layer to predict the phase:
\begin{equation*}
    [p_0, p_1] = \textrm{Softmax}\left(W^s\bm{h}^{\textrm{attr}} + \bm{b}^s\right).
\end{equation*}
Suppose we denote the predicted attribute by $a$. In the first part, we also use the same soft linear layer to predict it from content vocabulary:
\begin{equation*}
    p^{\textrm{new}}(a) = \textrm{Softmax}(W^{\textrm{new}}\bm{h}^{\textrm{attr}} + \bm{b}^{\textrm{new}}).
\end{equation*}
In the second part, different from \textit{instance}, we think the content comes from the token instead of the lemma of some word in sentence, so we have:
\begin{equation*}
    p_{1:m}^\textrm{token} = \textrm{Softmax}\left(\left(\bm{h}^{\textrm{attr}}\right)^\top\bm{t}_{1:m}\right),
\end{equation*}
where $\bm{t}_{1:m}$ are the embedding vectors of tokens which are introduced in Section~\ref{sec:sentencoder}.
The final probability of the predicted content $c$ is:
\begin{equation*}
    p(a) = p_0p^{\textrm{new}}(a) + p_1\sum_{i:w_i = a}p_i^{\textrm{token}}.
\end{equation*}

%If the next child is an attribute, similarly we will first convert $\bm{h}_f$ to attribute mode:
%\begin{equation*}
%    \bm{h}^{attribute} = \bm{h}_f + a\left(W^{attribute}\bm{h}_f\right).
%\end{equation*}
%We still divide the prediction of content into two part, new content part and copy part. The difference is, for the copy part, we use token not lemma for the copy mechanism, since attribute is usually a copy of the exactly word or phrase from sentence if it is a copy.

\subsection{Predict Edge}
The last thing we will predict is the edge (if $\bm{o} \not= 0$). We need to predict the \textit{label} part and the \textit{if-reverse} part. But before that, we need to produce a vector to represent the predicted vertex since we also need to use the vertex information here. There are two situations here:
\begin{itemize}
    \item \textbf{previously-produced vertex}: in this situation, we use the corresponding graph encoding vector to represent it. Suppose this previously-produced vertex is $v_i$, then we use 
    $\bm{g}_i^p$;
    \item \textbf{new instance or attribute}: in this situation, we use the method in Section~\ref{sec:vertex_embedding} to compute the vector.
\end{itemize}
Anyway, we use $\bm{v}^{\textrm{predict}}$ to denote this vector.

Now we are able to predict the edge. First, we predict the label part and we denote it by $e^l$:
\begin{equation*}
    p(e^l) = \textrm{Softmax}\left(W^{e,l}\left[\bm{h}^p; \bm{v}^{\textrm{predict}}\right] + \bm{b}^{e,l}\right).
\end{equation*}
Then, we use the edge embedding layer introduced in Section~\ref{sec:edge_embedding} to get the embedding vector $\bm{e}^l$ for it.

After that, we predict the \textit{if-reverse} part and we denote it by $e^r$:
\begin{equation*}
    p(e^r) = \textrm{Softmax}\left(W^{e,r}\left[\bm{h}^p; \bm{v}^{\textrm{predict}}; \bm{e}^{l}\right] + \bm{b}^{e,r}\right). 
\end{equation*}

\section{Hyper-Parameters} \label{sec:hyperparameter}
We use PyTorch for all the experiments. We try different sizes for the hidden vectors and embedding vectors from 128 to 1024 (128, 256, 512, 1024), and try different number of layers from 2 to 8. We use the Smatch score of the validation data to perform the hyperparameter selection. We show hyper-parameters of our best model in Table~\ref{tab:hyperparameter}. Here, \textit{graph hidden size} represents the size of all the embedding vectors (excluding the BERT vectors and refinement vectors, but they will finally be transformed to the \textit{graph hidden size}, see Section~\ref{sec:sentemb}), sentence encoding vectors, graph encoding vectors.

\begin{table}
    \centering
    \setlength\arrayrulewidth{1pt}
    \scalebox{1.0}{
    \begin{tabular} {|c|c|}
        \hline
        refinement embedding size & 300 \\
        \hline
        graph hidden size & 512 \\
        \hline
        graph encoding layers & 4 \\
        \hline
        sentence encoding layers & 4 \\
        \hline
        interactive Transformer layers & 4 \\
        \hline
        Transformer feed-forward hidden size & 1024 \\
        \hline
        number of Transformer heads & 8 \\
        \hline
        dropout & 0.1 \\
        \hline
        beam size & 8 \\
        \hline
    \end{tabular}
    }
    \caption{Hyper-parameters}
    \label{tab:hyperparameter}
\end{table}

\end{document}